\documentclass[sigconf,nonacm]{acmart}

\AtBeginDocument{%
  }

\acmConference[ICONS '24]{International Conference on Neuromorphic Systems}{July 30--Aug 2, 2024}{Arlington, VA}

\usepackage[]{acronym}
\usepackage{hyperref}
\usepackage{multirow}
\usepackage{graphicx}
\usepackage{subcaption}

\newcommand{\sd}{{$\Sigma\Delta$}}

\begin{document}

\acrodef{ANN}[ANN]{Artificial Neural Network}
\acrodef{RNN}[RNN]{Recurrent Neural Network}
\acrodef{RSNN}[RSNN]{Recurrent Spiking Neural Network}
\acrodef{SNN}[SNN]{Spiking Neural Network}
\acrodefplural{SNN}[SNNs]{Spiking Neural Networks}
\acrodef{LPF}[LPF]{Low-Pass Filter}

\title{Accurate Mapping of RNNs on Neuromorphic Hardware with Adaptive Spiking Neurons}

\author{Gauthier Boeshertz} 
\email{gauthier.boeshertz@gmail.com} 
\affiliation{ETH Zurich
\country{Switzerland}}

\author{Giacomo Indiveri}
\orcid{0000-0002-7109-1689}
\email{giacomo@ini.uzh.ch}
\affiliation{Institute of Neuroinformatics, University of Zurich and ETH Zurich \country{Switzerland}}

\author{Manu Nair*}
\orcid{0000-0002-0182-8358}
\email{manu@synthara.ai}
\affiliation{Synthara \country{Switzerland}}

\author{Alpha Renner*}
\orcid{0000-0002-0724-4169}
\email{a.renner@fz-juelich.de}
 \affiliation{Forschungszentrum Jülich
 \country{Germany}}

\def \authors{Gauthier Boeshertz, Giacomo Indiveri, Manu Nair, and Alpha Renner}
\renewcommand{\shortauthors}{Gauthier Boeshertz, Giacomo Indiveri, Manu Nair, and Alpha Renner}

 \begin{abstract}
   Thanks to their parallel and sparse activity features, recurrent neural networks (RNNs) are well-suited for hardware implementation in low-power neuromorphic hardware. 
   However, mapping rate-based RNNs to hardware-compatible spiking neural networks (SNNs) remains challenging.
   Here, we present a \sd-low-pass RNN (lpRNN): an RNN architecture employing an adaptive spiking neuron model that encodes signals using \sd-modulation and enables precise mapping. The \sd-neuron communicates analog values using spike timing, and the dynamics of the lpRNN are set to match typical timescales for processing natural signals, such as speech. Our approach integrates rate and temporal coding, offering a robust solution for the efficient and accurate conversion of RNNs to SNNs. We demonstrate the implementation of the lpRNN on Intel's neuromorphic research chip Loihi, achieving state-of-the-art classification results on audio benchmarks using 3-bit weights. These results call for a deeper investigation of recurrency and adaptation in event-based systems, which may lead to insights for edge computing applications where power-efficient real-time inference is required.

\end{abstract}

\thanks{\textsuperscript{*} equal contribution \\ \\
This paper was accepted at the IEEE/ACM International Conference on Neuromorphic Systems, July 30--Aug 2, 2024, Arlington, VA \\ \\
\copyright2024 IEEE. Personal use of this material is permitted. Permission from IEEE must be
obtained for all other uses, in any current or future media, including
reprinting/republishing this material for advertising or promotional purposes, creating new
collective works, for resale or redistribution to servers or lists, or reuse of any copyrighted
component of this work in other works.
}


\maketitle

\section{Introduction}

Neuromorphic computing~\cite{mead1990neuromorphic} aims to build computing systems using principles derived from neural systems of animal brains. Its common design features, geared toward improved efficiency compared to conventional architectures, are massive parallelism, near- or in-memory computation, and asynchronous communication using unary events (spikes).
Recurrently connected networks (RNN) that have a state that evolves over time make the best use of these principles~\cite{davies2021advancing}. After being initially sidelined by the Transformer's~\cite{vaswani2017attention} strong performance in natural language processing (NLP)~\cite{brown2020language}, RNNs are now experiencing a revival~\cite{gu2023mamba,orvieto2023resurrecting} due to their linear scaling with the number of inputs.

Despite methods like surrogate gradients~\cite{neftci2019surrogate,bittar2022surrogate} and training frameworks~\cite{eshraghian2023training} overcoming initial issues, translating \ac{ANN} progress to Spiking Neural Networks (SNNs) and neuromorphic hardware remains challenging~\cite{renner2021backpropagation}. Two main approaches exist:

1) Directly training SNNs inherently accounts for constraints like quantization and enables efficient spike timing codes~\cite{zenke2018superspike,neftci2019surrogate,shrestha2018slayer}.

2) ANN-SNN conversion converts trained ANNs to SNNs~\cite{eshraghian2023training}, which allows leveraging the most recent advances in ANNs, such as regularizers or network structures.

While conversion works well for feedforward convolutional networks in computer vision~\cite{ding2021optimal}, for RNNs, where small deviations accumulate, conversion approaches are rare~\cite{diehl2016conversion}.

In this work, we extend and validate an RNN conversion approach~\cite{nair2019mapping} on Intel's digital neuromorphic research chip Loihi~\cite{davies2018loihi}.
The approach solves the RNN conversion issue in two ways: 
First, the sigma-delta (\sd) spiking neuron model~\cite{yoon2016lif,zambrano2016fast,oconnor2017temporally,nair2019mapping,nair2019ultra}, analogous to adaptive linear integrate-and-fire (aLIF) models~\cite{brette2005adaptive,bohte2012efficient,nair2019ultra}, provides a precise way to map analog activations to spikes, allowing an approximation with a controlled error. The \sd-neuron combines aspects of rate and temporal coding, representing signals through spike timings rather than just firing rates. 
Second, we integrate the \sd-neurons into a low-pass recurrent neural network (lpRNN) architecture. The lpRNN's slower dynamics make RNN to SNN mapping easier and cheaper, matching typical timescales in speech and biomedical applications. 
Finally, we test the model in non-spiking and spiking simulation and on Intel's neuromorphic research chip Loihi on two benchmark datasets and set a new state-of-the-art for on-chip speech classification.
\section{Model}

\subsection{The \sd-spiking neuron model}
\label{sec:snn_model}

In this work, we adapt the aLIF/\sd-neuron \cite{bohte2012efficient,yoon2016lif,zambrano2016fast,oconnor2017temporally,nair2019mapping} and implement it on the Loihi \cite{davies2018loihi} neuromorphic chip. Due to its four state variables, the model requires two compartments on Loihi 1. \sd-neurons communicate analog signals by transmitting spikes only when the difference between an internal state and the input exceeds a threshold. It can be described on Loihi with the following discrete-time dynamics equations, adapted from the aLIF~\cite{nair2019mapping}:
\begin{align}
I_{mem}(t+1) &= I_{mem}(t) -\frac{I_{mem}(t)}{\tau_{mem}} + i - s  \label{eq:loihi_eqs_start}\\
s(t+1) &= s(t) -\frac{s(t)}{\tau_s} +  w_{fb} \delta_i \label{eq:loihi_eqs_s}\\
i(t+1) &= i(t) -\frac{i(t)}{\tau_i} + u(t)  \\
u(t+1) &= u(t) -\frac{u(t)}{\tau_u} + \sum_{n=1}^{N}{\delta_n W_n^{SNN}} 
\label{eq:loihi_eqs}
\end{align}

\begin{figure*}[!ht]
    \centering
    \subfloat[\label{fig:sd_blockdiag} ]{\includegraphics[width=0.45\linewidth]{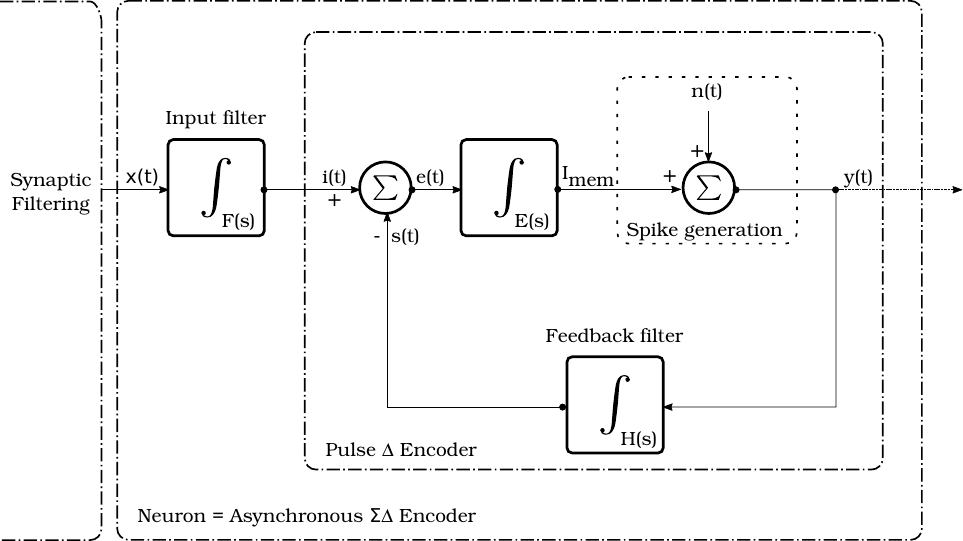}}
    \subfloat[\label{fig:sd_trans}]{\includegraphics[width=0.55\linewidth]{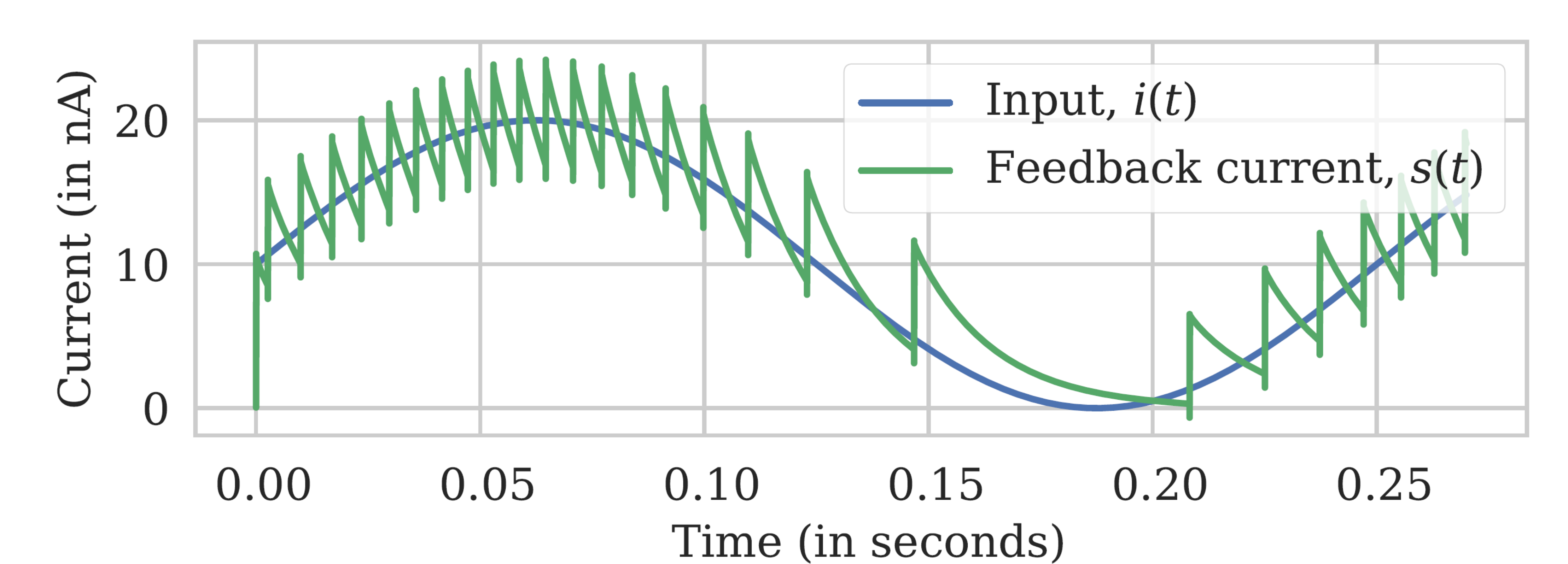}}
    \caption{\protect\subref{fig:sd_blockdiag}. Block diagram of a \sd-neuron.
    \protect\subref{fig:sd_trans}. {Evolution of the feedback signal $s(t)$ following the the input. Whenever it decays too far below the input, the neuron generates a spike ($y(t)$) which increases the feedback current. Panels from~\cite{nair2019mapping}.}}
    \label{fig:sd}
\end{figure*}

Where $I_{mem}$ represents the neuron's membrane potential, $s$ is the adaptation/feedback current, i is the input current, $b^{SNN}$ is the bias, u is the weighted (by $W_n^{SNN}$) and filtered input from presynaptic spike trains ($\delta_n$) or from an analog input signal (not shown), and $\tau_{mem},\tau_{s},\tau_{i},\tau_{u}$ are time constants. The neuron spikes when $I_{mem}$ exceeds a threshold, at which point $I_{mem}$ is reset to 0. 
This spike is transmitted to postsynaptic neurons and recursively added as a feedback spike train $\delta_i$ weighted by $w_{fb}$ to $s$.
This mechanism allows $s$ to track $i$ and serve as the neuron's internal activation state corresponding to an ANN activation value. Intuitively, whenever s diverges from i too much, a spike is produced to correct it (see Fig.~\ref{fig:sd_trans}).
This activation state can either be read out directly from $s$ or reconstructed by the output spikes. 
The model as a block diagram is shown in Fig.~\ref{fig:sd_blockdiag}. 

\begin{figure}
\centering
  \centering
\includegraphics[width=.5\linewidth]{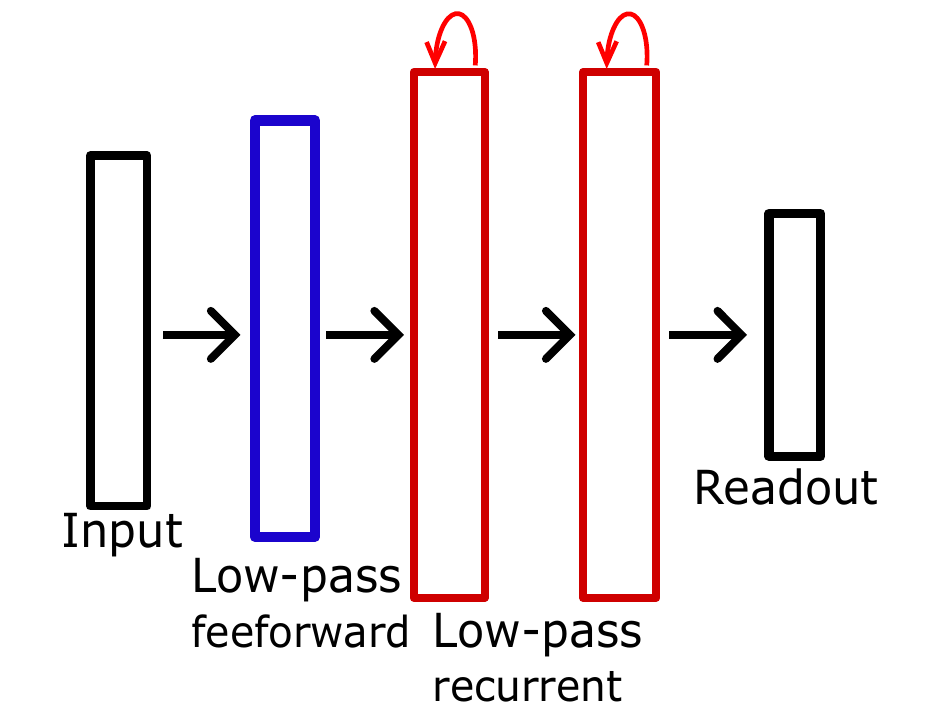}
  \caption{Architecture of the lpRNN network.}
  \label{fig:network_architecture}
\end{figure}

\subsection{The low-pass RNN model}
\label{sec:training}

The \sd-neuron model's feedback current $s$ corresponds to an ANN activation value $y$, enabling the mapping of RNNs to SNNs.
Moreover, the additional filtering of the input current $i$ implements a low-pass behavior, leading to a longer memory and, therefore, better performance than vanilla RNNs on specific tasks. Therefore, we term this the low-pass RNN (lpRNN), with non-spiking dynamics described by:
\begin{align}
\label{eq:lprnn}
y_t &= \alpha \odot y_{t-1} + (1-\alpha)\odot \sigma(W_{rec}\cdot y_{t-1} + W_{in} \cdot x_t + b)
\end{align}
where $\sigma$ is a non-linearity, $\odot$ the element-wise (Hadamard) product, and $\cdot$ matrix multiplication. $x_t$ the input vector, $y_t$ the output vector, the subscripts indicate the time step, $W_{rec}$ and $W_{in}$ the recurrent and input weight matrices, and $b$ the biases. Here, $\alpha$, a (reciprocal) time constant, is a fixed hyperparameter optimized by hyperparameter search but could be trained individually. $\alpha$ can be mapped to the time constants of the \ac{SNN} by $\tau = \frac{-T_s}{\log{\alpha}}$ where $T_s$ is the duration of the ANN algorithmic timestep or sampling interval of the input data-stream fed to the recurrent ANN. The nonlinearity $\sigma$ is a clamped ReLU as spiking neurons naturally implement a ReLU by not allowing a negative $s$~\cite{yoon2016lif,bohte2012efficient}. It is clamped to $I_{in}$ because the \ac{LPF} property of the \sd-neuron limits the maximum activity to $I_{in}$. 

While the ANN time resolution is given by the data sampling rate, the SNN requires a finer time resolution because the timing and the number of spikes matter. Therefore, the time constants are scaled by the ratio of the ANN and SNN algorithmic time steps, $\tau_{SNN} = \frac{T_{ANN}}{T_{SNN}} \tau_{ANN}$.
Due to constraints in the maximal neural state values on the Loihi chip, we use only 3 of the available 8-bit weight precision. Consequently, we train the ANN using 3-bit quantized weights in a quantization-aware manner via a straight-through estimator (STE) \cite{bengio2013estimating}.

The architecture used in this work, shown in Fig. \ref{fig:network_architecture}, consists of a feedforward low-pass layer to encode input signals into spikes, followed by two recurrent low-pass layers and an output layer. This compact size was chosen to avoid overfitting on the given datasets.
\section{Results}
We evaluate on two audio benchmark datasets commonly used for ANNs and SNNs: The Heidelberg Digits (HD) \cite{cramer2020heidelberg} and Google Speech Commands (GSC) \cite{warden2018speech}. For both datasets, we follow the same preprocessing, training, and testing procedure. The audio is transformed into Mel spectrograms, a common non-linear transformation \cite{deandrade2018neural,yin2021accurate,hammouamri2023learning} emphasizing lower frequencies where useful sounds reside \cite{davis1980comparison}.
We train the ANN weights and then transfer the network to an SNN for inference on the unseen test set. The input data is encoded into spikes using Brian2 \cite{stimberg2019brian}, simulating the input to the low-pass feedforward layer. The recurrent and output layers are run on Loihi and, as a baseline, in a Brian2 simulation. Further details are provided in the Methods and code (to be released).

\begin{table}[]
\caption{Top-1 Accuracy comparison on HD and SHD (20 classes).}
\label{table:shd_acc}

\begin{tabular}{lllll}
                     &       & Method & \begin{tabular}[c]{@{}l@{}}Weight\\ precision\\(bits)\end{tabular}& Acc. (\%)  \\ \midrule
\multirow{3}{*}{SHD} & SNN   & RadLIF  \cite{bittar2022surrogate}           & 32& 94.6          \\
                     &       & Learned Delays \cite{hammouamri2023learning} & 32& \textbf{95.1}          \\ \hline
\multirow{7}{*}{HD}  & ANN  & liBRU \cite{bittar2022surrogate}  & 32& \textbf{99.96} \\
                     &       & GRU \cite{bittar2022surrogate}  & 32 & 99.91        \\
                     &       & lpRNN [this work] & 3 & 99.69     \\
                     \cline{2-4} 
                     & SNN   & RLIF  \cite{bittar2022surrogate}  & 32 & 99.35     \\
                     &       & lpRNN [this work] & 3 & \textbf{99.69}         \\ \cline{2-4} 
                     & Loihi  & Speech2Spikes \cite{stewart2023speech2spikes} &  7 & 97.5     \\
                     & & lpRNN [this work] & 3 & \textbf{99.33} \\
                     \hline
\end{tabular}
\end{table}
The Heidelberg Digits (HD) \cite{cramer2020heidelberg} contains 10,000 recordings of 20 classes of digits in English and German. Instead of using the provided Spiking (SHD) version with spikes from an artificial cochlear model, we encode the raw audio Mel spectrograms using the \sd-neuron model. This approach produces fewer spikes (around 5600 vs. 8230 per test sample) while achieving better accuracy. As shown in Table \ref{table:shd_acc}, the lpRNN outperforms the only previous neuromorphic approach \cite{stewart2023speech2spikes} and matches state-of-the-art ANN and SNN methods. However, the near-perfect results indicate this dataset may not ideally differentiate model performance.

The Google Speech Commands (GSC) \cite{warden2018speech} contains one-second audio recordings of spoken words to be classified. We evaluate three variants: GSCv1 with 36 words and GSCv2 with 12 or 36 classes. To improve ANN-SNN mapping, we apply pruning \cite{frankle2019lottery} to reduce mismatch from low-activity neurons when $I_{in} >> activity$. Table \ref{table:gsc} compares results. The lower performance on GSCv2-36, compared to the best SNNs, is potentially due to using only 3-bit weights without batch normalization. The lpRNN again outperforms the previous neuromorphic approaches.

\setlength\tabcolsep{4pt}
\begin{table}[]
\caption{Top-1 accuracy comparison on the variants of the Google Speech Commands dataset}
\label{table:gsc}
\begin{tabular}{lllll}
&   \begin{tabular}[c]{@{}l@{}l@{}}Weight \\precision\\ (bits)\end{tabular} &v1-36w & v2-12w & v2-35w \\
\toprule
\textbf{Non-spiking}             &   &        &        &        \\
Att RNN \cite{deandrade2018neural}& 32 & 94.3   & 96.9   &  93.9  \\
AS Transformer \cite{gong2021ast} &  32&        &        &  98.1  \\
lpRNN ANN [this work]                & 3& 93.56  & 95.07  &  94.03      \\
\midrule
\textbf{Spiking}                   & 32&        &        &        \\
SRNN \cite{yin2021accurate}        & 32&        & 92.1   &        \\
LSNN \cite{salaj2021spike}         & 32&        & 91.2   &        \\
SLAYER \cite{orchard2021efficient} & 32&        & 91.74  &        \\
RLIF \cite{bittar2022surrogate}    & 32&        &        &  93.58 \\
RadLIF \cite{bittar2022surrogate}   & 32&       &        &  94.51 \\
Learned Delays \cite{hammouamri2023learning}&  32&        &  & 95.29 \\
lpRNN SNN [this work]            &   3& 93.19  & 93.13  &   93.33     \\
\midrule

\multicolumn{2}{l}{\textbf{Neuromorphic Hardware}}  &        &        \\
Speech2Spikes \cite{stewart2023speech2spikes}& 7&   &    &  71.1\\
Spinnaker2 Eprop \cite{rostami2022eprop} & 32&      & 91.12  &   \\
lpRNN Loihi [this work]       &  3& 92.08  & 92.8   &   92.4     \\

\bottomrule
\end{tabular}
\end{table}

\section{Discussion}

The lpRNN using \sd-neurons allows faithful conversion from \ac{RNN} to \ac{SNN}. We present one of the first results on a neuromorphic chip achieving comparable performance to \ac{RNN} and \ac{SNN} implemented on standard computing architectures. 
We demonstrate state-of-the-art results in the HD dataset in the SNN setting and strong results on GSC. This is achieved by training the network using backpropagation through time (BPTT) with the standard pipeline for ANNs and then transferring it to neuromorphic hardware with constraints on the quantization of weights (3-bit) and timesteps.

These results provide additional evidence that synaptic dynamics and neuronal adaptation are important mechanisms for encoding natural data with the \sd-mechanism. Most SNN methods in Table~\ref{table:gsc} use adaptive neuron models \cite{yin2021accurate,salaj2021spike,bittar2022surrogate}, generally outperforming non-adaptive models.
Adaption mechanisms that change the neuron spiking threshold, instead of subtracting an adaptation variable $s(t)$ from the membrane potential (Eq.\ref{eq:loihi_eqs_s}), are equivalent; they both change the amount of input needed until the threshold is reached. 

Compared to \ac{ANN} models, the lpRNN achieves results comparable to the commonly used gated RNNs (GRU and LSTM) on both datasets. This suggests that complex models with many parameters are not needed for time series without fast changes and without long-term dependencies, such as audio and biomedical data. 
However, for much longer sequences like sentences or documents, gated RNNs, state-space models, or transformers may be more suitable. Therefore, here, we focus on simple keyword spotting tasks, but the results encourage further exploration of the \sd-model, independently of the network architecture. The \sd-model is of particular interest for robust encoding and computation in ultra-low power analog hardware \cite{nair2019ultra, rubino2020ultra}.
Training the synaptic delays (as~\cite{hammouamri2023learning}) and time constants $\alpha$ individually might broaden the range of time scales of the network dynamics, making it suitable for additional applications.

\section{Methods}

\begin{figure}
\centering
  \centering
\includegraphics[width=.75\linewidth]{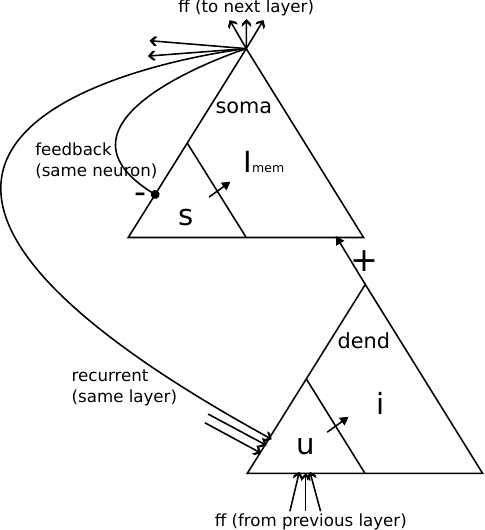}
  \caption{Multi-compartment \sd-neuron on Loihi}
  \label{fig:neuron_architecture}
\end{figure}

\label{sec:loihi_mapping}
The ANN simulations were made using Pytorch 1.9 
and the SNN simulations with Brian2~\cite{stimberg2019brian}.
Loihi experiments were conducted remotely using NxSDK version 0.9.9 on the Nahuku32 board ncl-ghrd-01. 

On Loihi, neurons can be created from one or more compartments. A compartment's input current can integrate spikes, and then this current is further integrated into the membrane potential. These signals can be relayed between compartments via connecting dendrites. The low-pass neuron comprises two compartments, as shown in Fig~\ref{fig:neuron_architecture}. The `dendritic' compartment integrates the spikes coming from recurrent and feedforward connections into the input current $u$. The `membrane potential' $i$ of this compartment is then added to the membrane potential $I_{mem}$ of the `somatic' compartment. This second compartment is connected to itself with an inhibitory synapse such that the spikes it generates are integrated into the adaptation/feedback current $s$.

The weights and biases for Loihi are obtained from the ANN as follows:
\begin{align}
    W^{SNN} = \frac{f W^{ANN}}{\tau_u*\tau_i*64} \qquad b^{SNN} = \frac{f b^{ANN}}{\tau_i}
\end{align}
A factor $f$ is used to scale the weights to make the best use of the precision available on the chip. Because the value of the state variables is bounded by $\pm2^{23}$ (24 bits), $f$ cannot be chosen too large to avoid overflow. A lower $f$, however, leads to lower weight precision, as weights are constrained to integer values. 

\begin{acks}
The authors thank Intel Labs for providing access to the Loihi research hardware. 
A.R. discloses support from the University of Zurich postdoc grant [FK-21-136] and the VolkswagenStiftung [CLAM 9C854].

\end{acks}

\bibliographystyle{ACM-Reference-Format}
\bibliography{references}

\end{document}